\documentclass[sigconf]{acmart}

\usepackage{booktabs} 

\setcopyright{rightsretained}

\acmDOI{10.475/123_4}

\acmISBN{123-4567-24-567/08/06}

\acmConference[GECCO '17]{the Genetic and Evolutionary Computation Conference 2017}{July 15--19, 2017}{Berlin, Germany}
\acmYear{2017}
\copyrightyear{2017}

\acmPrice{15.00}

\begin{document}
\title[Toward the automated analysis of complex diseases using genetic programming]{Toward the automated analysis of complex diseases in genome-wide association studies using genetic programming}

\author{Andrew Sohn}
\affiliation{
  \institution{University of Pennsylvania}
  \streetaddress{3700 Hamilton Walk}
  \city{Philadelphia} 
  \state{PA} 
  \postcode{19143}
}
\email{ansohn@upenn.edu}

\author{Randal S.~Olson}
\affiliation{
  \institution{University of Pennsylvania}
  \streetaddress{3700 Hamilton Walk}
  \city{Philadelphia} 
  \state{PA} 
  \postcode{19143}
}
\email{olsonran@upenn.edu}

\author{Jason H.~Moore}
\affiliation{
  \institution{University of Pennsylvania}
  \streetaddress{3700 Hamilton Walk}
  \city{Philadelphia} 
  \state{PA} 
  \postcode{19143}
}
\email{jhmoore@upenn.edu}

\renewcommand{\shortauthors}{A. Sohn et. al.}

\begin{abstract}

Machine learning has been gaining traction in recent years to meet the demand for tools that can efficiently analyze and make sense of the ever-growing databases of biomedical data in health care systems around the world. However, effectively using machine learning methods requires considerable domain expertise, which can be a barrier of entry for bioinformaticians new to computational data science methods. Therefore, off-the-shelf tools that make machine learning more accessible can prove invaluable for bioinformaticians. To this end, we have developed an open source pipeline optimization tool (TPOT-MDR) that uses genetic programming to automatically design machine learning pipelines for bioinformatics studies. In TPOT-MDR, we implement Multifactor Dimensionality Reduction (MDR) as a feature construction method for modeling higher-order feature interactions, and combine it with a new expert knowledge-guided feature selector for large biomedical data sets. We demonstrate TPOT-MDR's capabilities using a combination of simulated and real world data sets from human genetics and find that TPOT-MDR significantly outperforms modern machine learning methods such as logistic regression and eXtreme Gradient Boosting (XGBoost). We further analyze the best pipeline discovered by TPOT-MDR for a real world problem and highlight TPOT-MDR's ability to produce a high-accuracy solution that is also easily interpretable.

\end{abstract}

%
%


\maketitle

\section{Introduction}

We are currently witnessing the explosive growth of technologies that focus on processing the large amounts of data available in the biomedical sciences. Closely, in parallel, machine learning has been gaining traction in an effort toward analyzing and making sense of said biomedical data. However, effectively using machine learning tools often requires deep knowledge and expertise of both machine learning techniques as well as the application domain. For example, to effectively apply machine learning to a genome-wide association study (GWAS)~\cite{bird2007perceptions, cordell2009detecting}, the practitioner must understand the complex trait being studied (e.g., a particular disease such as prostate cancer), the research surrounding the underlying genetics of the trait, as well as the numerous steps in the machine learning process that are necessary for a successful analysis (e.g., data preprocessing, feature engineering, model selection, etc.). If we can provide off-the-shelf tools that reduce the barrier to entry for using machine learning by non-experts, then such tools could prove beneficial to researchers working in the biomedical sciences. Mapping statistical inferences and models from genetic data analysis to underlying biological processes is an important goal to the field of computational genomics~\cite{ma2002functional}.

In recent years, evolutionary computation (EC) has been proven successful in automating a variety of tasks, and even outperformed several hand-designed solutions in human vs. machine competitions~\cite{hornby2011computer,fredericks2013exploring,forrest2009genetic,spector2008genetic}. As such, we believe there is considerable promise in using EC to automate the analysis of biomedical data. Last year, we introduced the Tree-Based Pipeline Optimization Tool (TPOT)~\cite{Olson2016EvoBio,Olson2016GPTP}, which seeks to automate the process of designing machine learning pipelines using genetic programming (GP)~\cite{banzhaf1998genetic}. We found that TPOT often outperforms a standard machine learning analysis, all the while requiring no {\em a priori} knowledge about the problem it is solving~\cite{OlsonGECCO2016,Olson2016JMLR}. Here, we report on our attempts to specialize TPOT for human genetics research.

Human genetics research poses a unique data analysis challenge due to the effects of non-additive gene-gene interactions (i.e., epistasis) and the large number of genes that must be simultaneously considered as possible predictors of a complex trait~\cite{moore2010bioinformatics}. As a result, simple linear models of complex traits often predict little about the trait, and it is typically impossible to perform an exhaustive combinatorial search of every possible genetic model including two or more genes. For this reason, many researchers leverage {\em a priori} expert knowledge to intelligently reduce and guide the search space when performing a combinatorial search of possible genetic models~\cite{moore2006exploiting}.

In this paper we introduce TPOT-MDR, which uses GP to automate the study of complex diseases in GWAS. TPOT-MDR automatically designs sequences of common operations from genetic analysis studies, such as data filtering and Multifactor Dimensionality Reduction (MDR)~\cite{ritchie2001multifactor, hahn2003multifactor, moore2002new, cho2004multifactor, moore2006flexible, moore2015epistasis}, with the goal of producing a model that best predicts the outcome of a complex trait based solely on their genetics. Furthermore, we enable TPOT-MDR to leverage {\em a priori} expert knowledge through an Expert Knowledge Filter (EKF), which performs feature selection on the GWAS datasets using information from the expert knowledge source.

To demonstrate TPOT-MDR's capabilities, we compare TPOT-MDR to state-of-the-art machine learning methods on a combination of simulated and real-world GWAS datasets. These datasets are all supervised classification datasets with a focus on human disease as the outcome. We find that TPOT-MDR performs significantly better than the state-of-the-art machine learning methods on the GWAS datasets, especially when it is provided the EKF as an optional feature selector. We further analyze the resulting TPOT-MDR model on a real-world GWAS dataset to highlight the interpretability of TPOT-MDR models, which is a feature that is typically lacking in machine learning models. Finally, we release TPOT-MDR as an open source Python software package to be freely used in human genetics research.

\section{Related Work}

For automated machine learning in general, approaches have mainly focused on optimizing subsets of a machine learning pipeline~\cite{hutter2015beyond}, which is otherwise known as hyperparameter optimization. One readily accessible approach is grid search, which applies brute force search within a search space of all possible model parameters to find the best model configuration. Relatively recently, randomized search~\cite{bergstra2012random} and Bayesian optimization~\cite{snoek2012practical} techniques have entered into the foray and have offered more intelligently derived solutions---by adaptively choosing new configurations to train---to the hyperparameter optimization task. Much more recently, a novel bandit-based approach to hyperparameter optimization have outperformed state-of-the-art Bayesian optimization algorithms by 5x to more than an order of magnitude for various deep learning and kernel-based learning problems~\cite{li2016hyperband}. Although TPOT-MDR is an automated machine learning approach, it is more specialized on bioinformatics problems rather than general machine learning.

Narrowing the focus to automated machine learning in bioinformatics, the literature is far more sparse. One such example is~\cite{franken2012inferring}, in which they analyze metabolomics data using a modified Bayesian optimization algorithm integrated with the classification algorithms provided in WEKA, a suite of machine learning software written in Java. The Bayesian optimization provided feature subset selection, which filtered irrelevant and redundant features from the datasets to achieve dimensionality reduction. These techniques lead to an improvement of classification accuracy.

Genetic programming and evolutionary computation methods have also been successfully applied to bioinformatics studies, such as~\cite{Moore2013,urbanowicz2013role}, but they do not focus on designing and tuning a series of standard data analysis operations for a specific dataset. As such, although they are related techniques, they do not fall into the automated machine learning domain.

\section{Methods}

In this section, we briefly review TPOT~\cite{Olson2016EvoBio,OlsonGECCO2016,Olson2016JMLR,Olson2016GPTP} and describe the new pipeline operators that were implemented for TPOT-MDR. Afterwards, we describe the datasets used to evaluate TPOT-MDR and compare it to the state-of-the-art machine learning methods.

\subsection{TPOT Review}
\label{sec:tpot-review}

TPOT uses an evolutionary algorithm to automatically design and optimize a series of standard machine learning operations (i.e., a pipeline) that maximize the final classifier's accuracy on a supervised classification dataset. It achieves this task using a combination of genetic programming (GP)~\cite{banzhaf1998genetic} and Pareto optimization (specifically, NSGA2~\cite{Deb2002}), which optimizes over the trade-off between the number of operations in the pipeline and the accuracy achieved by the pipeline.

TPOT implements four main types of pipeline operators: (1) preprocessors, (2) decomposition, (3) feature selection, and finally (4) models. All the pipeline operators make use of existing implementations in the Python scikit-learn library~\cite{pedregosa2011scikit}. Preprocessors consist of two scaling operators to scale the features and an operator that generates new features via polynomial combinations of numerical features. Decomposition consists of a variant of the principal component analysis (\texttt{RandomizedPCA}). Feature selection implements various strategies that serve to filter down the features by some criteria, such as the linear correlation between the feature and the outcome. Models consist of supervised machine learning models, such as tree-based methods, probabilistic and non-probabilistic models, and k-nearest neighbors. 

TPOT combines all the operators described above and assembles machine learning pipelines from them. When a pipeline is evaluated, the entire dataset is passed through the pipeline operations in a sequential manner---scaling the data, performing feature selection, generating predictions from the features, etc.---until the final pipeline operation is reached. Once the dataset has fully traversed the pipeline, the final predictions are used to evaluate the overall classification accuracy of the pipeline. This accuracy score is used as part of the pipeline's fitness criteria in the GP algorithm.

To automatically generate and optimize these machine learning pipelines, TPOT uses a GP algorithm as implemented in DEAP~\cite{fortin2012deap}, which is a Python package for evolutionary algorithms. Oftentimes, GP builds trees of mathematical functions that seek to optimize toward a specified criteria. In TPOT, GP is used to optimize the number and order of pipeline operators as well as each operator's parameters. TPOT follows a standard GP process for 100 generations: random initialization of the initial population (default population size of 100), evaluation of the population on a supervised classification dataset, selection of the most fit individuals on the Pareto front via NSGA2, and variation through uniform mutation (90\% of all individuals per generation) and one-point crossover (5\% of all individuals per generation). For more information on the TPOT optimization process, see~\cite{OlsonGECCO2016}.

\subsection{TPOT-MDR}

\begin{figure*}
    \includegraphics[width=\textwidth]{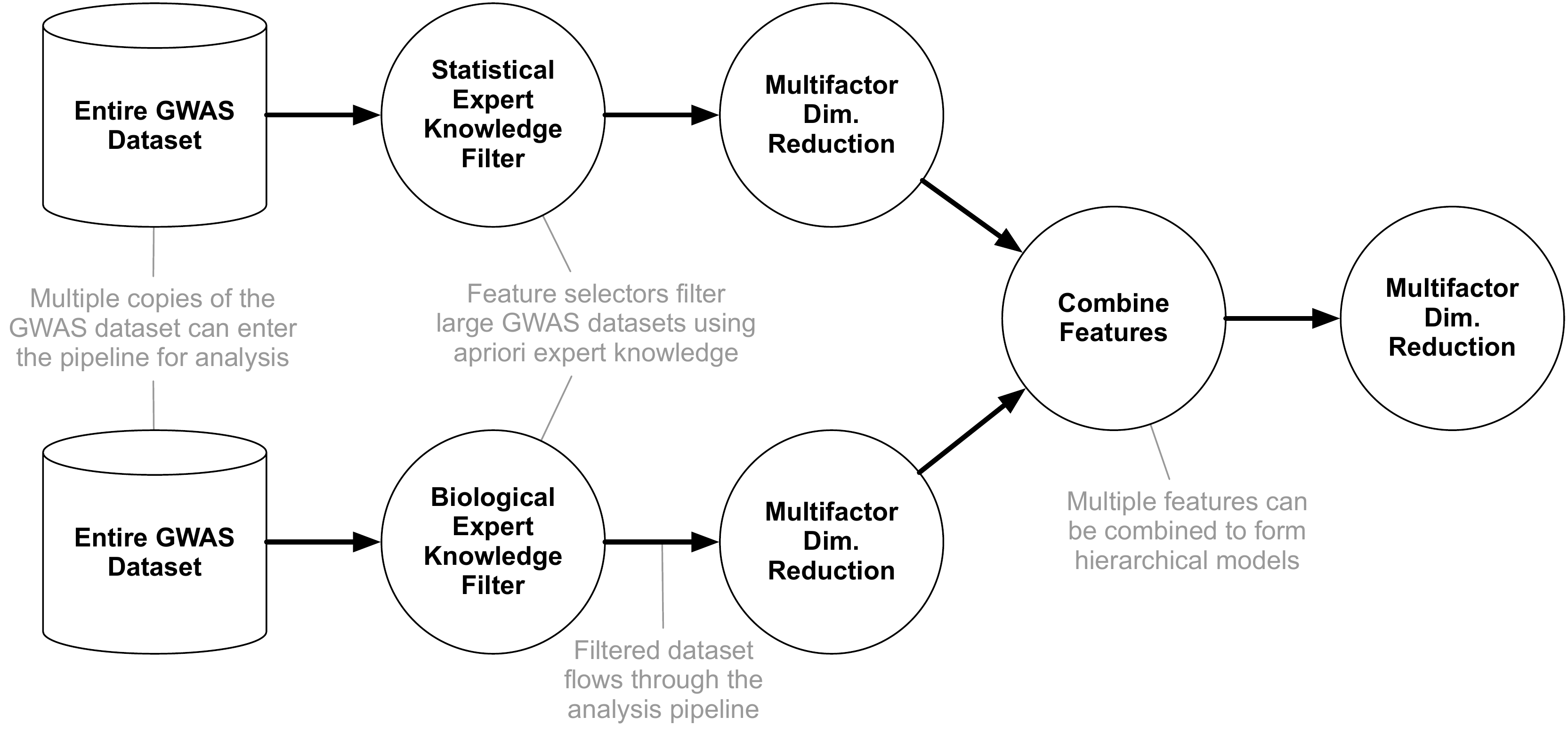}
    \centering
    \caption{Example TPOT-MDR pipeline. Each circle represents an operation on the dataset, and each arrow represents the passing of the processed dataset to another operation.}
    \label{fig:tpot-mdr-example}
\end{figure*}

TPOT-MDR is a specialized version of TPOT that focuses on genetic analysis studies. It features two new operators that are commonly used genetic analyses of human disease: (1) Multifactor Dimensionality Reduction (MDR) and (2) an Expert Knowledge Filter (EKF).

MDR is a machine learning method for detecting statistical patterns of epistasis by manipulating the feature space of the dataset to more easily identify interactions within the data~\cite{ritchie2001multifactor, hahn2003multifactor, moore2006flexible, moore2015epistasis}. To summarize, MDR is a constructive induction algorithm that combines two or more features to create a single feature that captures the interaction affects among the features. This constructed created feature can be fed back into the dataset as a new feature or used as the final prediction on the dataset.

The motivation behind adding the EKF operator was that, often times, {\em a priori} expert knowledge about a biomedical dataset exists: Perhaps the dataset has been analyzed and annotated in previous studies, a database exists with relevant information about the genes in a dataset, or statistical expert knowledge can be derived from the dataset before the study~\cite{moore2010bioinformatics}. This {\em a priori} expert knowledge can be leveraged to guide the TPOT-MDR search algorithm in deciding what genes to include in the final genetic model.

The EKF operator selects an expert knowledge source from the sources provided and selects the \texttt{N} best features according to the expert knowledge source (where \texttt{N} is constrained to [1, 5]). Since the EKF operator is parameterized to select both the expert knowledge source and the number of top features to retain, TPOT-MDR optimizes (1) whether and where in the pipeline to include the EKF and (2) the parameters of the EKF. Multiple EKF operators can be included in a TPOT-MDR pipeline, as shown in Figure~\ref{fig:tpot-mdr-example}.

Other than the MDR and EKF operators, the only other operators included in TPOT-MDR are a standard univariate feature selection method (\texttt{SelectKBest} in scikit-learn~\cite{pedregosa2011scikit}, with an evolvable number of features to retain, \texttt{N}, where \texttt{N} is constrained to [1, 5]) and a \texttt{CombineDFs} operator that combines two feature sets together into a single feature set. These operators can be chained together to form a series of operations acting on a GWAS dataset, as depicted in Figure~\ref{fig:tpot-mdr-example}. Except for different operator set, the TPOT-MDR optimization process works the same as the original TPOT algorithm as described in Section~\ref{sec:tpot-review}, and was run with a population size of 300 for 300 generations with a per-individual mutation rate of 90\% and per-individual crossover rate of 5\%.

\subsection{Datasets}

We performed an analysis of TPOT-MDR on both simulated datasets and a real world GWAS dataset. The simulated datasets were generated using GAMETES~\cite{urbanowicz2012gametes}, an open source software package designed to generate GWAS datasets with pure epistatic interactions between the features. We simulated 16 different datasets with specific properties to test the scalability of TPOT-MDR. The simulated datasets included 10, 100, 1,000, or 5,000 single-nucleotide polymorphism (SNP) features, each with 2 predictive features and the remaining features generated randomly using an allele frequency between 0.05 and 0.5. Further, we generated datasets with heritabilities (i.e., noise) of 0.05, 0.1, 0.2, or 0.4, where lower heritability entails more noise in the dataset. Notably, all of the GAMETES datasets had a sample size of 2,000 to ensure a reasonably large dataset size.

By scaling the GAMETES dataset feature spaces from 10 to 5,000, we sought to evaluate how well TPOT-MDR could handle increasingly large numbers of non-predictive features. Similarly, by simulating increasing amounts of noise in the dataset, we sought to evaluate how much noise TPOT-MDR could handle before it failed to detect and model the predictive features. As such, this simulated benchmark provides a detailed view of of the strengths and limitations of TPOT-MDR in the GWAS domain.

To validate TPOT-MDR on a real-world dataset, we used a nationally available genetic dataset of 2,286 men of European descent (488 non-aggressive and 687 aggressive cases, 1,111 controls) collected through the Prostate, Lung, Colon, and Ovarian (PLCO) Cancer Screening Trial, a randomized, well-designed, multi-center investigation sponsored and coordinated by the National Cancer Institute (NCI) and their Cancer Genetic Markers of Susceptibility (CGEMS) program. In this study, we focus on prostate cancer aggressiveness as the endpoint, where the prostate cancer is considered aggressive if it was assigned a Gleason score $\geq$ 7 and was in tumor stages III/IV. Between 1993 and 2001, the PLCO Trial recruited men ages 55--74 years to evaluate the effect of screening on disease specific mortality, relative to standard care. All participants signed informed consent documents approved by both the NCI and local institutional review boards. Access to clinical and background data collected through examinations and questionnaires was approved for use by the PLCO. Men were included in the current analysis if they had a baseline PSA measurement before October 1, 2003, completed a baseline questionnaire, returned at least one Annual Study Update (ASU), and had available SNP profile data through the CGEMS data portal\footnote{http://cgems.cancer.gov}. Prior to this study, the CGEMS dataset was filtered to the 219 SNPs associated with biological pathways relevant to aggressive prostate cancer~\cite{Lavender2012Interaction}. We call this dataset the ``CGEMS Prostate Cancer GWAS dataset.''

For all experiments, we used four different statistical expert knowledge sources as input to the EKF operator: the ReliefF~\cite{kononenko1997overcoming}, SURF~\cite{greene2009spatially}, SURF*~\cite{greene2010informative}, and MultiSURF~\cite{granizo2013multiple} algorithms. These algorithms evaluated the entire dataset prior to the experiments and assigned numerical feature importance scores to each feature, which is an indication of how predictive each feature is of the outcome. These numerical scores were provided to the TPOT-MDR EKF operator, and were used to rank the features when filtering the datasets. We computed the statistical expert knowledge sources for all 16 GAMETES datasets and the CGEMS Prostate Cancer GWAS dataset, resulting in 68 unique expert knowledge sources (4 for each experiment).

\subsection{Evaluating TPOT-MDR}
\label{sec:evaluating-tpot-mdr}

We ran four different sets of experiments on the datasets: (1) Extreme Gradient Boosting (XGBoost)\footnote{XGBoost parameters: 500 trees, learning rate 0.0001, and 10 maximum tree depth}~\cite{chen2016xgboost}, (2) Logistic Regression\footnote{The logistic regression regularization parameter was tuned via 10-fold cross validation}~\cite{MachineLearningBook}, (3) TPOT-MDR without the EKF, and (4) TPOT-MDR with the EKF. In Section~\ref{sec:results}, we refer to these experiments as \texttt{XGBoost}, \texttt{Logistic Regression}, \texttt{TPOT (MDR only)}, and \texttt{TPOT (MDR + EKF)}, respectively. For the GAMETES datasets, we additionally compared the four experiments to the baseline of a MDR model constructed with the two known predictive SNP features (called \texttt{MDR (Predictive SNPs)}), which will achieve the maximum possible classification accuracy for the GAMETES datasets without overfitting on the noisy features.

We chose to compare TPOT-MDR to the XGBoost classifier because XGBoost has been established as a widely popular and successful tree-based classifier in the machine learning community, particularly in the Kaggle\footnote{http://www.kaggle.com} machine learning competitions. Further, we compared TPOT-MDR to a logistic regression to demonstrate the capabilities of a standard linear model on GWAS datasets, which will essentially detect only linear associations between the features and the outcome. Finally, we ran TPOT-MDR without the EKF to demonstrate whether the EKF was important for the TPOT-MDR optimization process.

For every dataset and experiment, we performed 30 replicate runs with unique random number seeds (where applicable). This allowed us to evaluate and explore the limits of TPOT-MDR's modeling capabilities on a broad range of GWAS datasets, and demonstrate how it performs in comparison to state-of-the-art machine learning methods. In all cases, the accuracy scores reported are averaged balanced accuracy scores from 10-fold cross-validation, where the balanced accuracy metric is a normalized version of accuracy that accounts for class imbalance by calculating accuracy on a per-class basis then averaging the per-class accuracies~\cite{Velez2007,urbanowicz2015exstracs}. With balanced accuracy, a score of 50\% is equivalent to random guessing, even with imbalanced datasets.

\section{Results}
\label{sec:results}

\begin{figure*}
    \includegraphics[width=\textwidth]{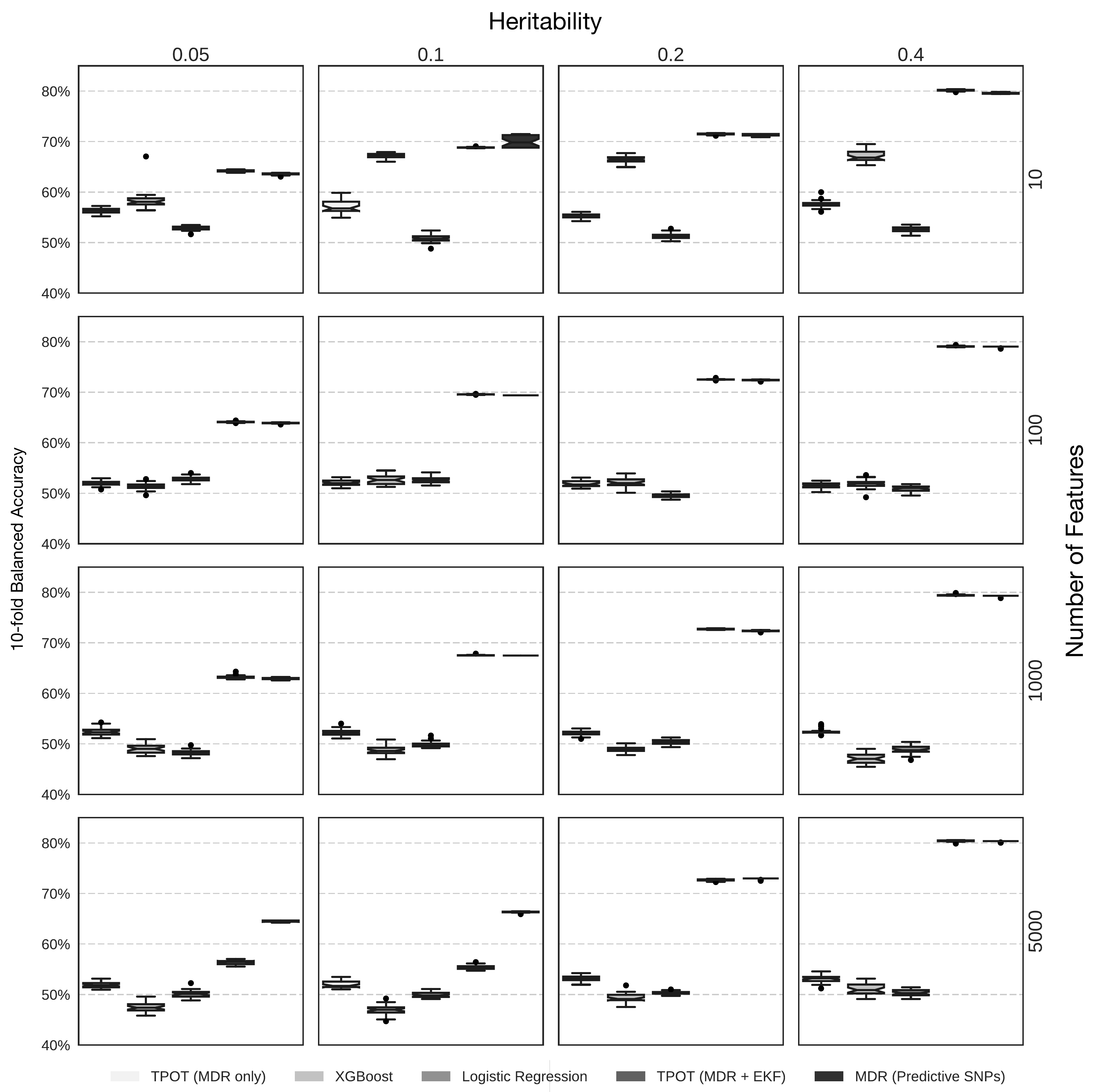}
    \centering
    \caption{Comparison of results on the simulated GAMETES GWAS datasets. Each box plot shows the distribution of averaged 10-fold balanced accuracies for each experiment, where the notches indicate the 95\% confidence interval. A 50\% balanced accuracy is equivalent to random guessing. Each panel within the figure corresponds to differing levels of heritability (i.e., dataset noise) and numbers of features in the simulated datasets, ranging from the easiest dataset on the top right (high heritability, small numbers of features) to the hardest dataset bottom left (low heritability, large numbers of features).\\\\Since some of the experiments had little variance in scores, some box plots are too small to determine their color. For clarity, the box plots represent the following experiments, in order from left to right: TPOT (MDR only), XGBoost, Logistic Regression, TPOT (MDR + EKF), and MDR (Predictive SNPs). These experiments are described in Section~\ref{sec:evaluating-tpot-mdr}.}
    \label{fig:gametes-comparison}
\end{figure*}

\begin{figure*}
    \includegraphics[width=0.4\textwidth]{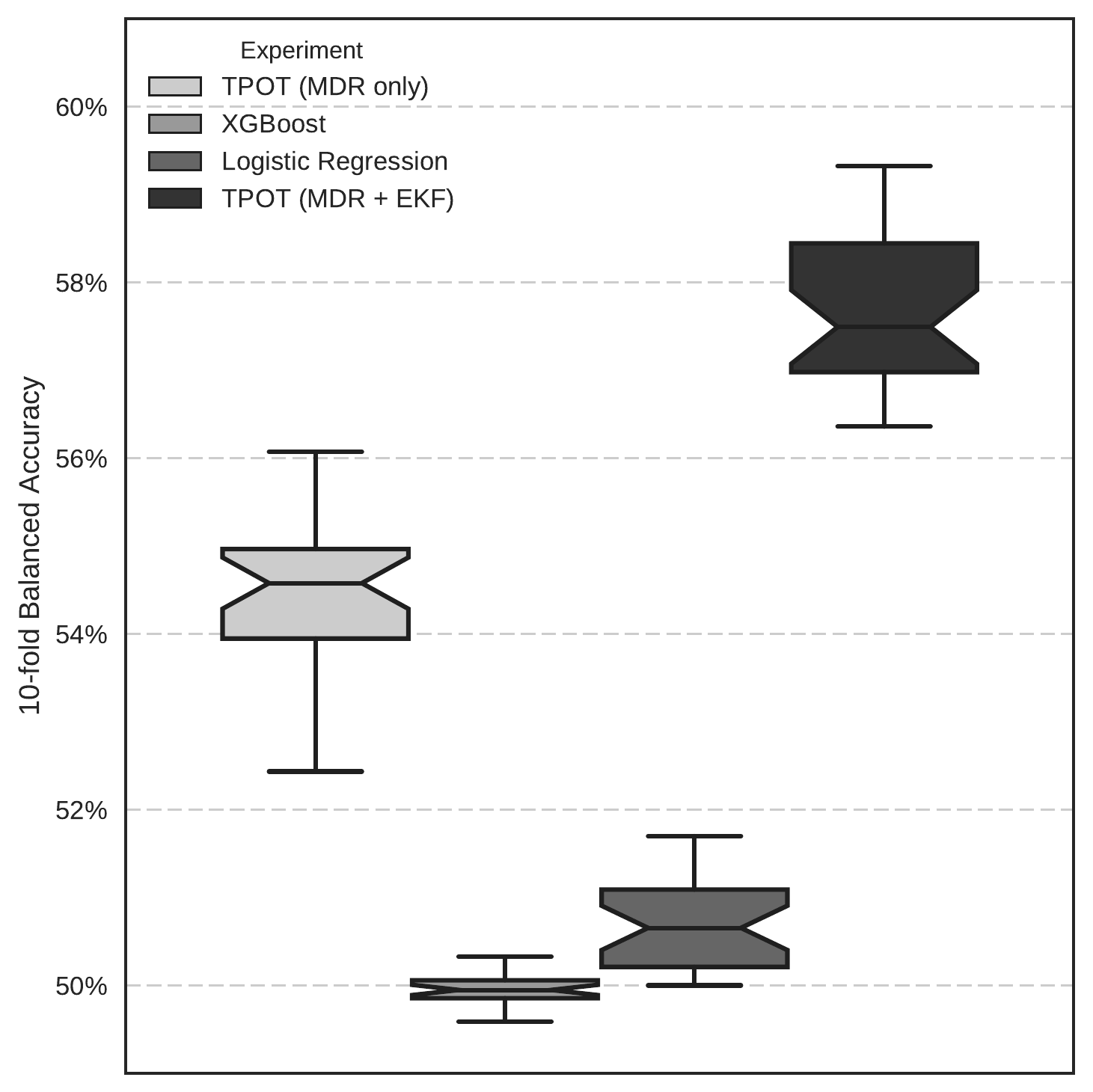}
    \centering
    \caption{Comparison of results on the CGEMS prostate cancer GWAS dataset. Each box plot shows the distribution of averaged 10-fold balanced accuracies for each experiment, where the notches indicate the 95\% confidence interval. A 50\% balanced accuracy is equivalent to random guessing.}
    \label{fig:cgems-comparison}
\end{figure*}

\begin{figure*}
    \includegraphics[width=\textwidth]{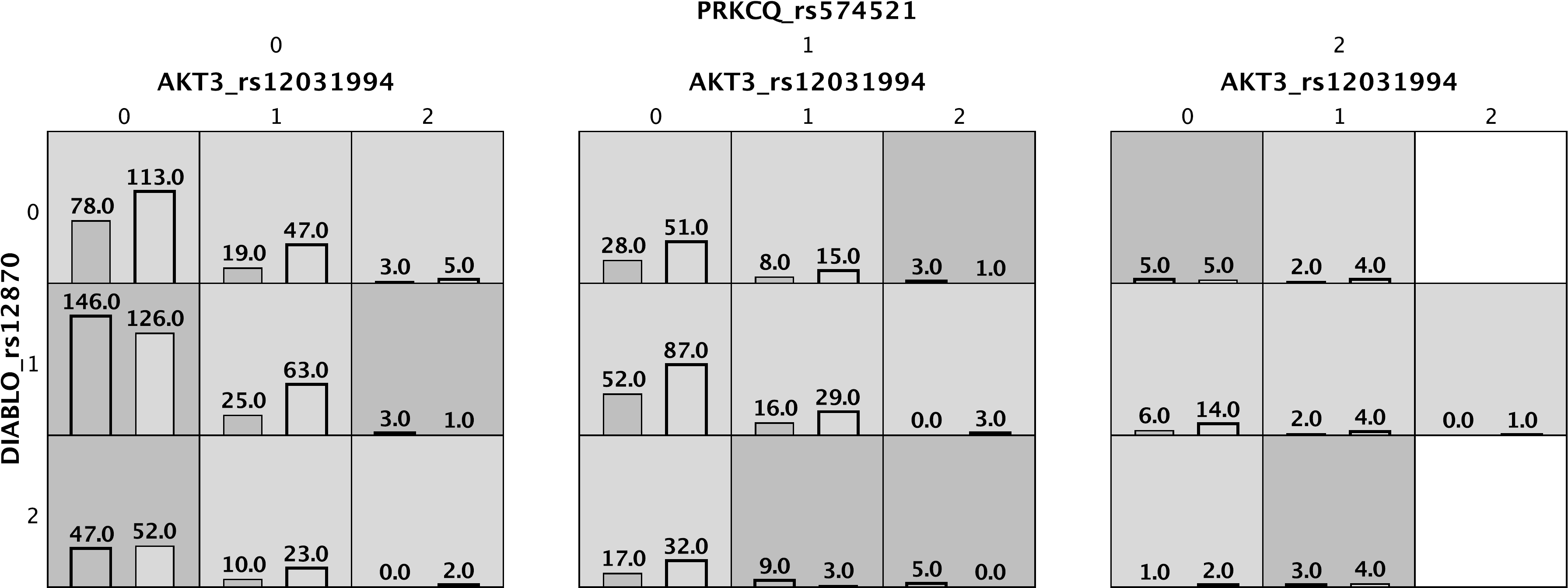}
    \centering
    \caption{Classification grid for the best MDR model that TPOT-MDR discovered for the CGEMS prostate cancer GWAS dataset. Each of the three grids correspond to one state of the \texttt{PRKCQ\_rs574512} SNP, whereas the cells within each grid correspond to one combination of states between the \texttt{AKT3\_rs12031994} and \texttt{DIABLO\_rs12870} SNPs. Thus, for example, the light grey upper right cell in the leftmost grid corresponds to \texttt{PRKCQ\_rs574512} = 0, \texttt{AKT3\_rs12031994} = 2, and \texttt{DIABLO\_rs12870} = 0.\\\\Dark grey bars and cells indicate aggressive cases (i.e., at risk of aggressive prostate cancer), whereas light grey bars and cells indicate non-aggressive cases (i.e., lower risk of aggressive prostate cancer). The numbers at the top of each bar indicate the number of aggressive and non-aggressive cases that fall within each cell when the entire CGEMS dataset is sorted into the MDR classification grid. If no data points fall into a cell, the cell is left blank.}
    \label{fig:cgems-mdr-grid}
\end{figure*}

\subsection{GAMETES Simulated Datasets}

As shown in Figure~\ref{fig:gametes-comparison}, TPOT-MDR without the EKF rarely finds the best genetic model because it only has a univariate feature selector at its disposal. In contrast, TPOT-MDR with the EKF always discovers the best genetic model except when there are thousands of features and high noise. Even in the cases where TPOT-MDR with the EKF fails to find the best genetic model, it still discovers better genetic models than the other methods in this study.

For a baseline, we compared TPOT-MDR to a tuned logistic regression and XGBoost, as described in Section~\ref{sec:evaluating-tpot-mdr}. Figure~\ref{fig:gametes-comparison} shows that logistic regression consistently fails to find a good model and barely performs better than chance in even the easiest GAMETES datasets. This finding demonstrates a key flaw in using linear models for GWAS: Linear models will not detect higher-order interactions within the dataset unless the interactions are explicitly modeled. Similarly, XGBoost can sometimes find a good model for GWAS datasets if the dataset is heavily filtered beforehand (e.g., to 10s of features), but rapidly degrades in performance as more noisy features are added to the dataset.

\subsection{CGEMS Prostate Cancer Dataset}

The CGEMS prostate cancer GWAS dataset has 219 SNPs, 1,175 samples, and likely falls into the ``lower heritability'' spectrum of the GAMETES datasets. Thus, we would expect to see roughly similar performance on the CGEMS dataset as we saw in the GAMETES datasets with 100 features and 0.1 or 0.05 heritability in Figure~\ref{fig:gametes-comparison}.

As predicted, Figure~\ref{fig:cgems-comparison} shows that XGBoost and logistic regression fail to discover the higher-order interactions within the real-world CGEMS dataset. In contrast, TPOT-MDR with and without the EKF managed to consistently find predictive genetic models for the CGEMS dataset. In particular, TPOT-MDR with the EKF found the best genetic models, largely because the expert knowledge sources (ReliefF, SURF, etc.) contained information about the higher-order interactions between the SNPs that TPOT-MDR was able to harness.

To better understand the genetic models that TPOT-MDR discovered, we analyzed the final model from the highest-scoring TPOT-MDR experiment and visualized the pattern of interactions from the MDR model in Figure~\ref{fig:cgems-mdr-grid}. We see patterns suggestive of statistical epistasis within the model, for example, in the leftmost grid a patient's aggressive (dark grey cells) or non-aggressive (light grey cells) status can only be determined by a combination of \texttt{AKT3\_rs12031994} and \texttt{DIABLO\_rs12870}. Similarly, the pattern of aggressive vs. non-aggressive status between \texttt{AKT3\_rs12031994} and \texttt{DIABLO\_rs12870} varies depending on the state of the third SNP, \texttt{PRKCQ\_rs574512}, which suggests a statistical three-way epistatic interaction between the SNPs. If there were no higher-order interactions between the SNPs, then we would expect a patient's aggressive vs. non-aggressive status to vary independently between the SNPs, i.e., we would expect to see horizontal and vertical bands of aggressive or non-aggressive status within the grids. As previous studies have suggested links between these SNPs and aggressive prostate cancer~\cite{Lavender2012Interaction}, we can use these TPOT-MDR findings to further elucidate the SNPs' higher-order interactions and involvement in the development of aggressive prostate cancer in men of European descent.

\section{Discussion}

In this paper, we introduced a new method and tool, TPOT-MDR, for automating the analysis of complex diseases in genome-wide association studies (GWAS). We developed this tool to aid bioinformaticians so they can more efficiently process and analyze the ever-growing databases of biomedical data. To that end, TPOT-MDR is designed to optimize a series of machine learning operations that are commonly used in biomedical studies, such as filtering the features using expert knowledge sources, combining information from different expert knowledge sources, and modeling the higher-order interactions of the features using Multifactor Dimensionality Reduction (MDR) to predict a patient's outcome. Before, bioinformaticians would typically perform and refine these operations by hand, whereas now TPOT-MDR can relieve the bioinformatician of these tedious duties so they can focus on more challenging tasks.

Even though this paper focuses on the application of TPOT-MDR to GWAS datasets, we note that TPOT-MDR is a general machine learning tool that will work with any dataset that has categorical features and a binary outcome. TPOT-MDR has been released as a free, open source Python tool and is available on GitHub\footnote{https://github.com/rhiever/tpot/tree/tpot-mdr}.

In Section~\ref{sec:results}, we evaluated TPOT-MDR on a series of simulated and real-world GWAS datasets and found that TPOT-MDR outperforms linear models and XGBoost across all of the datasets (Figures~\ref{fig:gametes-comparison} and~\ref{fig:cgems-comparison}). These findings are important for several reasons. For one, we demonstrated that simple linear models are ill-suited for the analysis of GWAS datasets owing to their inability to model higher-order interactions within the dataset. We also demonstrated that state-of-the-art tree-based machine learning methods---typically thought to be effective at modeling higher-order feature interactions---are similarly ill-suited for modeling GWAS datasets with large numbers of features. Finally, we highlighted the importance of harnessing {\em a priori} expert knowledge to filter GWAS datasets prior to the modeling step, which could aid state-of-the-art machine learning algorithms such as XGBoost in eliminating extraneous features.

Although the results in Section~\ref{sec:results} suggest that TPOT-MDR is superior to the compared methods on every dataset we used, there are some drawbacks to TPOT-MDR that must be considered. For one, linear models and XGBoost are orders of magnitude faster to train and evaluate than TPOT-MDR. As TPOT-MDR uses genetic programming to optimize the series of filtering and modeling operations on the dataset, a single TPOT-MDR run took roughly 3 hours on the CGEMS dataset, whereas XGBoost and logistic regression each took less than a minute. Given that many GWAS datasets often have thousands to hundreds of thousands of SNP features (compared to the 219 in CGEMS), TPOT-MDR will require more work to improve its run time scalability to larger GWAS datasets. Furthermore, TPOT-MDR is highly dependent on its expert knowledge sources. In these experiments, we used expert knowledge sources that specialize in detecting higher-order epistatic interactions, which proved to be critical in both the simulated and real world datasets. If TPOT-MDR is provided with less informative expert knowledge sources, then it will likely perform worse, which we can observe in Figures~\ref{fig:gametes-comparison} and~\ref{fig:cgems-comparison} (TPOT-MDR without EKF vs. TPOT-MDR with EKF).

As shown in Figure~\ref{fig:gametes-comparison}, XGBoost can sometimes model higher-order interactions when the dataset is heavily filtered beforehand. However, the resulting XGBoost model is not nearly as interpretable as with TPOT-MDR. TPOT-MDR produces a model that we can inspect to study the pattern of feature interactions within the dataset (Figure~\ref{fig:cgems-mdr-grid}), whereas XGBoost provides only a complex ensemble of decision trees. This is an important consideration when building machine learning tools for bioinformatics: More often than not, bioinformaticians do not need a black box model that achieves high prediction accuracy on a real-world dataset. Instead, bioinformaticians seek to build a model that can be used as a microscope for understanding the underlying biology of the system they are modeling. In this regard, the models generated by TPOT-MDR can be invaluable for elucidating the higher-order interactions that are often present in complex biological systems.

In conclusion, TPOT-MDR is a promising step forward in using evolutionary algorithms to automate the design of machine learning workflows for bioinformaticians. We believe that evolutionary algorithms (EAs) are poised to excel in the automated machine learning domain, and specialized tools such as TPOT-MDR highlight the strengths of EAs by showing how easily EA solution representations can be adapted to a particular domain.

\section{Acknowledgements}

We thank the Penn Medicine Academic Computing Services for the use of their computing resources. This work was supported by National Institutes of Health grant AI116794.

\newpage

\bibliographystyle{ACM-Reference-Format}
\bibliography{references} 

\end{document}